\documentclass{isprs} 
\usepackage{subfigure}
\usepackage{setspace}
\usepackage{geometry} 
\usepackage{epstopdf}
\usepackage[labelsep=period]{caption}  
\usepackage[british]{babel} 
\usepackage[hang]{footmisc}
\usepackage{amsmath}
\usepackage[most]{tcolorbox}

\begin{document}

\title{Zero-shot Vision-Language Reranking for Cross-View Geolocalization}
\date{}

 \author{
  Yunus Talha Erzurumlu \textsuperscript{1}, John E. Anderson\textsuperscript{2}, William J. Shuart\textsuperscript{2}, Charles  Toth\textsuperscript{3}, Alper Yilmaz\textsuperscript{3} }
\address{
	\textsuperscript{1 }Dept. of Electrical and Computer Engineering, The Ohio State University,  281 W Lane Ave, Columbus, Ohio, USA \\ erzurumlu.1@osu.edu\\
	\textsuperscript{2 }US Army Corps of Engineers Geospatial Research Lab, Corbin Field Station, Woodford, Virginia, USA \\
    (john.anderson, william.j.shuart)@erdc.dren.mil\\
    \textsuperscript{3 }Dept. of Civil Engineering, The Ohio State University, 281 W Lane Ave, Columbus, Ohio, USA \\ (toth.2, yilmaz.15)@osu.edu\\
}



\abstract{
Cross-view geolocalization (CVGL) systems, while effective at retrieving a list of relevant candidates (high Recall@k), often fail to identify the single best match (low Top-1 accuracy). This work investigates the use of zero-shot Vision-Language Models (VLMs) as rerankers to address this gap. We propose a two-stage framework: state-of-the-art (SOTA) retrieval followed by VLM reranking. We systematically compare two strategies: (1) \textbf{Pointwise} (scoring candidates individually) and (2) \textbf{Pairwise} (comparing candidates relatively). Experiments on the VIGOR dataset show a clear divergence: all pointwise methods cause a catastrophic drop in performance or no change at all. In contrast, a \textbf{pairwise comparison} strategy using LLaVA \textbf{improves Top-1 accuracy} over the strong retrieval baseline. Our analysis concludes that, these VLMs are poorly calibrated for \textit{absolute} relevance scoring but are effective at fine-grained \textit{relative} visual judgment, making pairwise reranking a promising direction for enhancing CVGL precision.
}

\maketitle

\section{Introduction}
\label{introduction}

\begin{figure}[ht!]
\begin{center}
		\includegraphics[width=1.0\columnwidth]{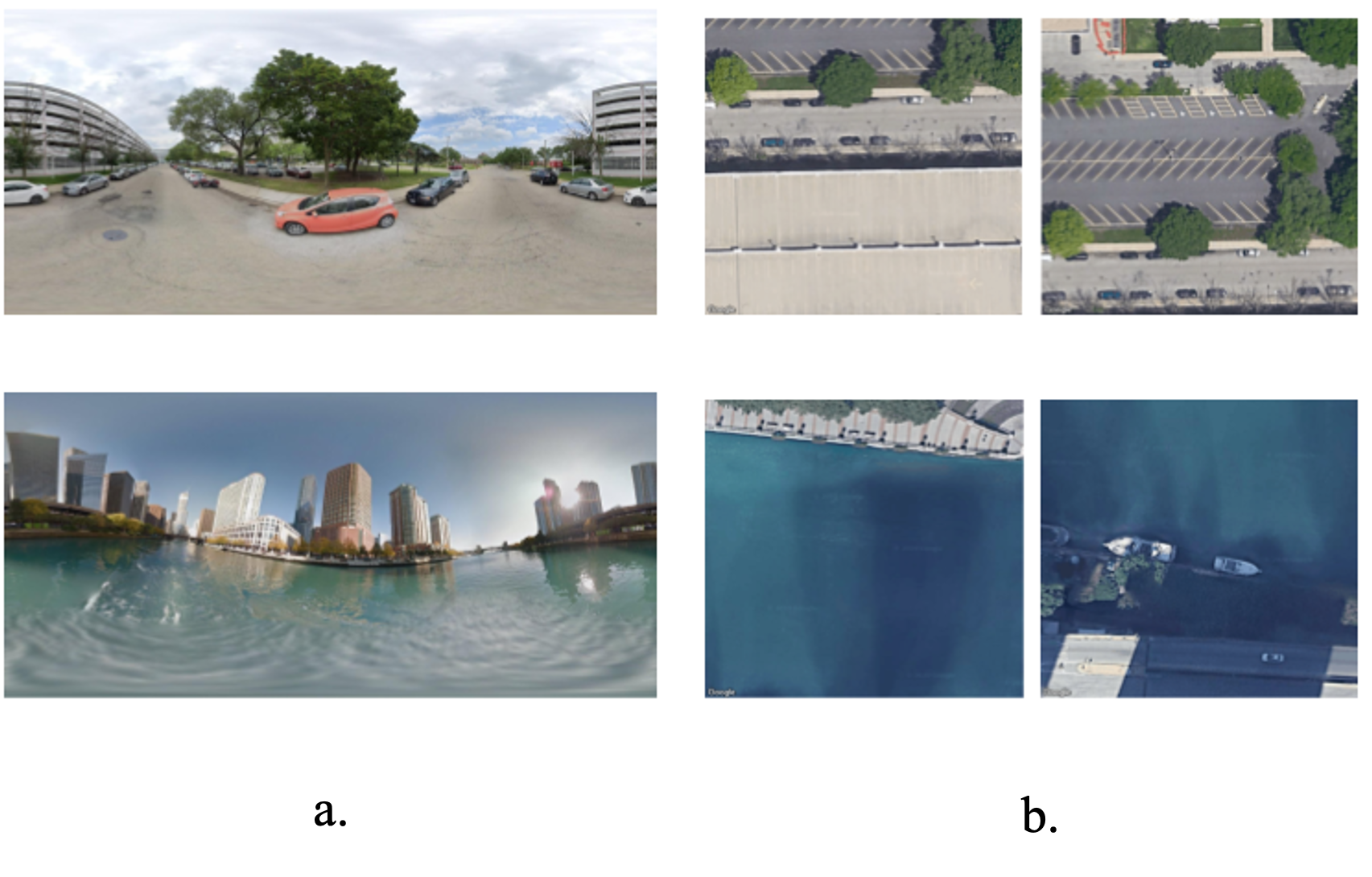}
	\caption{Example of the cross-view geolocalization challenge. a. Ground-level query images. b. Several aerial image candidates, only one of which is the correct match, demonstrating the difficulty caused by viewpoint and appearance differences.}
\label{fig:CVGL}
\end{center}
\end{figure}

Cross-view geolocalization (CVGL), the task of determining the geographic location of a ground-level query image by matching it against a database of geo-referenced aerial or satellite images, is a fundamental challenge in computer vision with significant real-world applications. These include autonomous vehicle and drone navigation, augmented reality systems, and location-based services \cite{Lin2014CVGL,Workman2015WideArea}. The core difficulty lies in bridging the drastic viewpoint and appearance gap between ground and aerial perspectives, often compounded by variations in season, illumination, and structural changes over time \cite{Zhu2020VIGORCI}. Figure \ref{fig:CVGL} illustrates this challenge.

Current state-of-the-art (SOTA) methods, often based on contrastive image retrieval, have achieved impressive results in retrieving a set of relevant candidates, demonstrating high recall rates at top-k positions (e.g., Recall@10, Recall@20) \cite{Deuser2023Sample4GeoHN}. However, they frequently struggle with identifying the single best match, resulting in relatively low top-1 accuracy. This limitation hinders deployment in scenarios requiring high precision. These models typically rely on learning global image representations that capture coarse-grained similarities but may overlook the fine-grained details and semantic understanding necessary for exact matching.

\begin{figure}[ht!] 
  \centering
  \includegraphics[width=1\columnwidth]{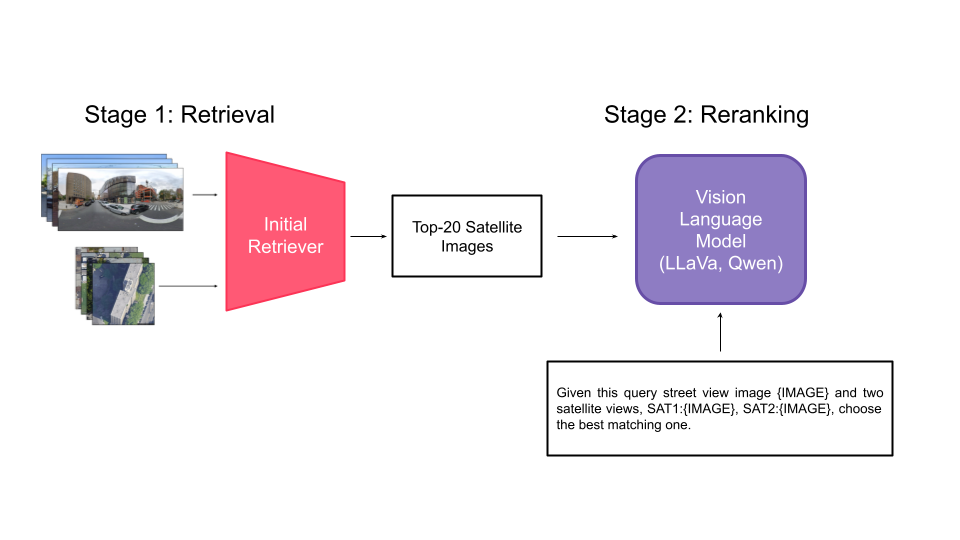} 
  \caption{The proposed two-stage framework. Stage 1 uses a SOTA retrieval model to get top-$K$ candidates . Stage 2 uses a VLM reranker to produce the final ranked list.}
  \label{fig:framework}
\end{figure}

Recently, Vision-Language Models (VLMs) such as LLaVA \cite{Liu2023VisualIT} and Qwen-VL \cite{Bai2025Qwen25VLTR} have demonstrated remarkable abilities in multimodal understanding and reasoning. These models can process and interpret information from both visual and textual modalities, enabling complex tasks like visual question answering, image captioning, and detailed region description. Inspired by the success of Large Language Models (LLMs) in reranking tasks within text retrieval (e.g., RankGPT \cite{Sun2023IsCG}), we hypothesize that VLMs can leverage their visual-textual reasoning capabilities to perform fine-grained reranking for cross-view geolocalization.
In this work, we propose a two-stage framework (visualized in Figure \ref{fig:framework}) to improve top-1 geolocalization accuracy. First, a SOTA cross-view retrieval model generates an initial list of top-20 candidate aerial images for a given ground query. Second, we employ a VLM to rerank these candidates. We systematically investigate different VLM prompting strategies for reranking:
\begin{itemize}
    \item \textbf{Pointwise}: Evaluating each aerial candidate independently against the ground query using direct score prediction, binary (Yes/No) relevance prediction, or Likert scale ratings. We also explore variants incorporating explicit reasoning prompts.
    \item \textbf{Pairwise}: Comparing two aerial candidates at a time to determine which better matches the ground query, subsequently using these relative judgments to sort the full list.
\end{itemize}
We evaluate these strategies using LLaVA-1.5-7b and Qwen2.5VL-7b on the VIGOR dataset \cite{Zhu2020VIGORCI}. Our key finding is that the pairwise comparison strategy, when implemented with LLaVA, outperforms both pointwise methods and the strong retrieval baseline, achieving a top-1 accuracy of 64.80\% compared to the baseline 61.20\%. This suggests that VLMs could be more effective at making relative fine-grained visual judgments than assigning absolute relevance scores in this challenging cross-view setting. 

Importantly, our focus is deliberately on the zero-shot setting: rather than adapting VLMs to the task through additional training, we ask whether off-the-shelf VLMs already possess enough cross-view reasoning ability to improve retrieval outputs through prompting alone.

\section{Related Works}
\label{related_work}

\subsection{Cross-View Geolocalization}
CVGL problem initially introduced by \cite{Lin2014CVGL} and \cite{Workman2015WideArea}.
Recent advances in the problem have been dominated by deep learning. Convolutional Neural Networks (CNNs) \cite{Workman2015WideArea,Shi2019SpatialAwareFA,Zhu2020VIGORCI} and later Transformers \cite{Zhu2022TransGeoTI,Zhu2023SAIG} have been employed to learn robust feature representations invariant to viewpoint changes. Architectures like Siamese networks \cite{Koch2015SiameseNN} are commonly used to learn a shared embedding space where ground and aerial images of the same location are close, often trained with contrastive or triplet loss functions \cite{Oord2018RepresentationLW,Hoffer2014DeepML}. Attention mechanisms \cite{Zhu2022TransGeoTI} and feature aggregation techniques \cite{Arandjelovic2015NetVLAD,Hu2018CVMNet,Shi2019SpatialAwareFA,Zhu2023SAIG} have further improved performance by focusing on salient regions. While these methods achieve high recall@k $>$ 5, recall@1 still remains a challenge \cite{Zhu2020VIGORCI}. Furthermore, reasoning-based frameworks such as GeoReasoner \cite{Li2024GeoReasonerGW} leverage street-view reasoning to refine candidate selection, and alignment-tuning approaches like AddressVLM \cite{xu2025addressvlm} enable fine-grained address-level localization. Zero-shot and open-domain methods such as StreetCLIP \cite{Haas2023LearningGZ} reduce dependence on large paired datasets by transferring generalized embeddings across diverse geographic settings. Our work focuses not on improving the initial retrieval but on reranking the outputs of such SOTA models to take advantage of already strong recall@k $>$ 5 metrics.

\subsection{Vision-Language Models (VLMs)}
VLMs combine vision encoders like ViT \cite{Dosovitskiy2020AnII} with LLMs like Llama \cite{Touvron2023LLaMAOA} to perform joint visual and textual understanding. Models like CLIP \cite{Radford2021LearningTV} pioneered aligning images and text in a shared embedding space. More recent models like LLaVA \cite{Liu2023VisualIT} and Qwen-VL \cite{Bai2025Qwen25VLTR} adopt an instruction-following paradigm, enabling them to perform diverse multimodal tasks based on natural language prompts by connecting a vision encoder to an LLM via a projection layer and fine-tuning on instruction datasets. Their ability to understand spatial relationships, object attributes, and scene semantics alongside textual reasoning makes them promising candidates for the fine-grained analysis required in geolocalization reranking.

\subsection{LLMs/VLMs for Reranking}
The idea of using large generative models for reranking has gained traction, primarily in text retrieval. RankGPT \cite{Sun2023IsCG} and similar approaches \cite{Chen2024AttentionIL} demonstrated that LLMs can effectively rerank document lists using pointwise, pairwise, or listwise prompting strategies, often surpassing traditional learning-to-rank methods. Extending this to the visual domain is less explored. Our work applies this reranking concept using VLMs for the specific, challenging task of cross-view geolocalization, focusing on comparing different prompting strategies within this multimodal context.

\section{Methodology}
\label{methodology}

We propose a two-stage framework for cross-view geolocalization, leveraging a SOTA retrieval model for candidate generation and a VLM for fine-grained reranking, as illustrated in Figure \ref{fig:framework}.

\subsection{Stage 1: Candidate Retrieval}
Given a ground-level query image $I_g$, we first employ a pre-trained SOTA cross-view geolocalization model (e.g., based on \cite{Deuser2023Sample4GeoHN}) to retrieve the top-$K$ most similar aerial/satellite images $\{I_{a,1}, I_{a,2}, ..., I_{a,K}\}$ from a large, geo referenced database.

In our experiments, we use $K=20$. This choice is motivated by the strength of the first-stage retriever: the baseline model already achieves Recall@20 above 90\%, meaning that the correct match is typically present in the candidate pool while keeping VLM inference computationally tractable. Our goal in this work is therefore not to perform exhaustive VLM-based search over the full satellite database, which would be prohibitively expensive, but to study whether a zero-shot VLM can improve precision within a high-recall candidate set. This also highlights a limitation of the proposed framework: if the correct match is absent from the initial top-$K$ list, the reranker cannot recover it. Let this initial ranked list be $L_{initial}$.

\subsection{Stage 2: VLM Reranking}
The core of our approach lies in using a VLM to rerank the initial candidate list $L_{initial}$ to produce a refined list $L_{reranked}$, aiming to place the true match at the top rank. We utilize VLMs capable of processing both the ground query image $I_g$ and each aerial candidate image $I_{a,i}$. We explored several prompting strategies:

\subsubsection{Pointwise Strategies} These methods evaluate each aerial candidate $I_{a,i}$ independently with respect to the ground query $I_g$.
\begin{itemize}
    \item \textbf{Direct Prediction:} The VLM is prompted to output a score on 0-100 scale. Example prompt: "Assess the similarity... Score:" We directly take this as final score and rank candidates by this score. 
    \item \textbf{Likert Score:} The VLM is prompted to output a similarity score on a Likert scale (1-5). Example prompt: "Assess the similarity... Score:" We calculate the expected score based on the VLM's output probability distribution over the tokens '1', '2', '3', '4', '5' and rank candidates by this score. Given the VLM’s output probability $p_{i,k}=P(s=k\mid I_g,I_{a,i})$ over the tokens $k\in\{1,2,3,4,5\}$ for aerial candidate $i$, the expected score is:
    \[
    \hat{s}_i \;=\; \sum_{k=1}^{5} k\,p_{i,k}\,.
    \]

    \item \textbf{Yes/No Prediction:} The VLM is prompted for a binary relevance judgment. Example prompt: "Do the ground-level image and the aerial image show the same location?... Answer:" Candidates are ranked based on the probability assigned to the 'Yes' token, calculated as:
    \[\frac{P(\text{'Yes'})}{P(\text{'Yes'}) + P(\text{'No'})}\]
    \item \textbf{Reasoning + Yes/No:} We first ask the VLM to reason about the match before providing a Yes/No answer. Example prompt: "Compare the ground-level and aerial images... Reasoning and Answer:" Ranking is based on the final $P(\text{'Yes'})$ as above.
\end{itemize}

\subsubsection{Pairwise Strategies}
These methods compare two aerial candidates, $I_{a,i}$ and $I_{a,j}$, at a time against the ground query $I_g$.

\begin{itemize}
    \item \textbf{Pairwise Comparison:} The VLM is prompted to choose which of two aerial candidates is the better match for the ground query. In other words, for a fixed query $I_g$, the VLM acts as a comparison function $C(I_g, I_{a,i}, I_{a,j})$ that returns the preferred candidate.
\end{itemize}

We instantiate reranking as a comparison-based sorting procedure over the top-$K$ candidate list, where the VLM provides the outcome of each pairwise comparison and the sorting algorithm determines the final order. In our implementation, we use \textit{merge sort}, requiring approximately $O(K \log K)$ pairwise VLM calls.

We chose this formulation because our goal is to produce a full reranked list rather than only identify a single winner. While a tournament-style elimination scheme would require only $K-1$ comparisons to select the top candidate, it does not directly provide a reliable ordering of the remaining candidates, which is important for evaluating Recall@1, Recall@3, and Recall@5. In addition, full sorting is less dependent on a single early decision. Exploring lower-budget pairwise schemes such as elimination brackets is an interesting direction for future work.

\subsubsection{VLMs Used} We experimented with two publicly available VLMs: LLaVA-1.5-7b \cite{Liu2023VisualIT} and Qwen2.5VL-7b (referred to as Qwen-VL in the paper \cite{Bai2025Qwen25VLTR}).

\section{Experiments}
\label{experiments}

\subsection{Dataset and Setup}
We evaluate our approach on the VIGOR dataset \cite{Zhu2020VIGORCI}. VIGOR is a widely used benchmark for cross-view geolocalization and a challenging one. We follow standard protocols and use the provided test split for CROSS-AREA task,  focusing on matching visible ground queries to visible satellite images. We randomly sampled 500 queries for efficiency. The initial top-20 candidates ($K=20$) for each query were generated using a SOTA retrieval model Sample4Geo \cite{Deuser2023Sample4GeoHN}, providing a strong baseline.

\subsection{Evaluation Metrics}
We report standard retrieval metrics: Top-1 Accuracy (Recall@1), Recall@3, and Recall@5. Top-1 Accuracy is our primary metric, reflecting the ability to pinpoint the exact match. Recall@k measures the percentage of queries for which the correct aerial image is ranked within the top $k$ positions.
\begin{table}[ht!] 
  \caption{Performance comparison of VLM reranking strategies on the VIGOR dataset (500 test queries). Baseline performance is from the initial SOTA retrieval model. Results are shown in Recall@k (\%). '-' indicates metric not applicable or computed. Best result for each metric is in \textbf{bold}.}
  \label{tab:main_results}
  \centering
  \small 
  
  \begin{tabular}{|l|c|c|c|} 
    \hline 
    Method                     & R@1 (\%) & R@3 (\%) & R@5 (\%) \\
    \hline 
    Baseline (Retrieval Only)  & 61.20        & 73.80        & 82.40        \\
    \hline 
    \multicolumn{4}{|l|}{\textit{Pointwise Methods}} \\ 
    \hline 
        LLaVA Direct               & 61.20         & 73.80        & 82.40        \\
    Qwen Direct                & 25.00        & 49.00        & 65.20        \\
    LLaVA Likert               & 4.80         & 15.20        & 26.20        \\
    Qwen Likert                & 14.80        & 34.80        & 50.80        \\
    LLaVA Yes/No               & 15.20        & 33.40        & 46.00        \\
    Qwen Yes/No                & 15.60        & 36.20        & 53.20        \\
    LLaVA Reason Yes/No        & 8.00         & 21.60        & 34.00        \\
    Qwen Reason Yes/No         & 12.80        & 31.00        & 42.20        \\
    \hline 
    \multicolumn{4}{|l|}{\textit{Pairwise Methods}} \\
    \hline 
    LLaVA Pairwise             & \textbf{64.80}& \textbf{84.80}& \textbf{89.80}   \\
    Qwen Pairwise              & 30.40        & 55.00        & 68.40        \\

    \hline 
  \end{tabular}
\end{table}
\subsection{Implementation Details}
Experiments were run using the official implementations and pre-trained weights for LLaVA-1.5-7b and Qwen2.5VL-7b. We utilized NVIDIA A6000 GPUs for the VLM inference. 
For pairwise sorting, we implemented an efficient sorting algorithm using the VLM's comparison output. Probabilities for Yes/No and Likert scoring were extracted from the VLM's logits for the target tokens.

\begin{figure*}[ht!] 
  \centering
  \includegraphics[width=1\linewidth]{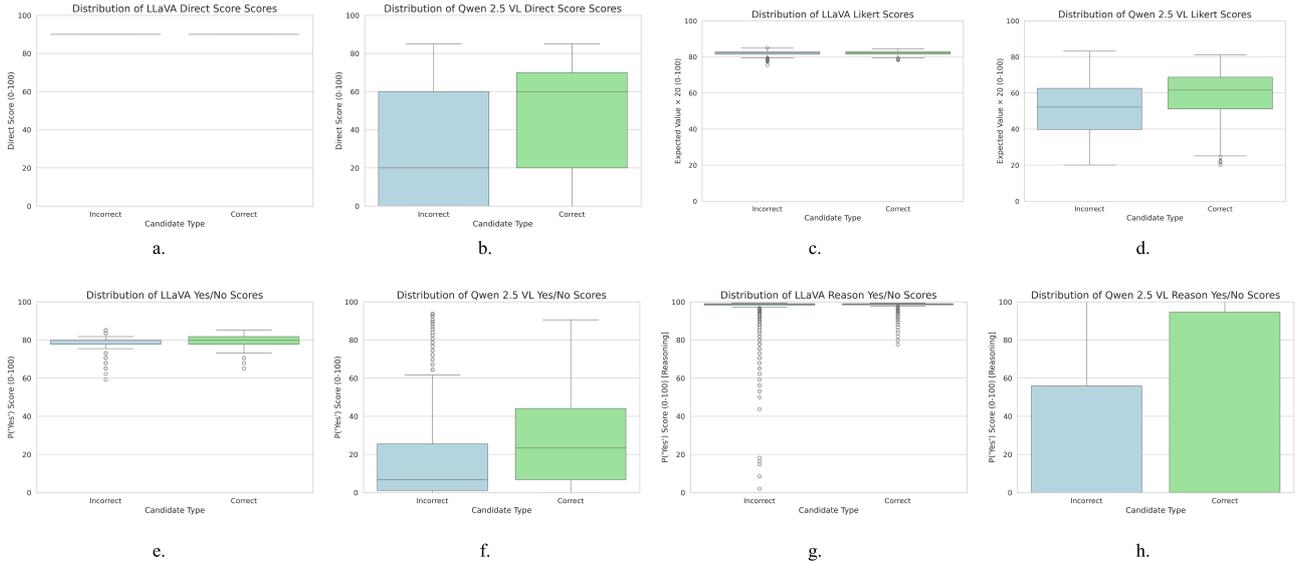} 
\caption{Analysis of pointwise reranking failures. Plots (a–b) show score distributions for direct score prediction, (c–d) for Likert-scale prediction, (e–f) for Yes/No prediction, and (g–h) for Yes/No prediction with explicit reasoning. In all cases, the distributions for correct and incorrect candidates strongly overlap, indicating that the VLM-assigned scores (whether direct, Likert, or Yes/No) do not provide a clear, separable signal to distinguish the true match from other plausible-but-incorrect candidates.}

  \label{fig:point_plots}
\end{figure*}

\subsection{The Ineffectiveness of Pointwise Reranking}

Our results demonstrate a clear and consistent failure of all pointwise reranking strategies. As shown in \textbf{Table \ref{tab:main_results}}, every pointwise method from Likert scales to binary Yes/No judgments performs significantly worse than the retrieval baseline, with Top-1 accuracies collapsing to a range between 4.80\% (LLaVA Likert) and 15.60\% (Qwen Yes/No).

The score distribution plots in \textbf{Figure \ref{fig:point_plots}} provide a clear explanation for this failure: the these VLMs appear  miscalibrated for assigning absolute relevance scores in this challenging cross-view task. The models exhibit two distinct failure modes:

\begin{itemize}
    \item \textbf{LLaVA: Lack of Discrimination.} LLaVA consistently struggles to produce a useful signal. Across all pointwise experiments (Likert, Yes/No, etc.), its score distributions for correct and incorrect candidates are nearly identical. This is taken to an extreme in the direct scoring method (Figure \ref{fig:point_plots}.a), where LLaVA assigns the same high score to virtually all candidates, resulting in a ranking identical to the baseline and yielding no change.

    \item \textbf{Qwen-VL: Signal Drowned by Variance.} In contrast, Qwen-VL \textit{does} show a clear tendency to assign higher average scores to correct matches (Figure \ref{fig:point_plots}.b). However, this potential signal is rendered ineffective by an extremely high variance. Many incorrect candidates also receive high scores, making it impossible to reliably rank the true match at the top.
\end{itemize}

Furthermore, we found that adding an explicit reasoning step (Reason Yes/No) consistently \textit{degraded} performance for both models. For Qwen, this prompt exacerbated the variance issue, pushing scores for all candidates even higher and further confusing the ranking. This collective failure strongly suggests that while VLMs can identify plausible candidates, they are unsuited for making the fine-grained, absolute similarity judgments required by pointwise reranking, without further training at least.

\subsection{The Success of Pairwise Comparison}

In a drastic departure from the pointwise failures, the pairwise comparison strategy yielded significantly different results. The LLaVA Pairwise method achieved the best performance overall, with a Top-1 accuracy of 64.80\%. This represents a 3.6\% absolute improvement over the strong 61.20\% baseline. The Recall@3 (84.80\%) and Recall@5 (89.80\%) metrics also improved, indicating a more robust ranking in the top-k positions.

However, this modest 3.6\% improvement from LLaVA must be interpreted with caution. Given LLaVA's total inability to discriminate between correct and incorrect candidates in \textit{any} pointwise setting (as shown in Figure \ref{fig:point_plots}), its success here may be limited. It is plausible that the model is only resolving "easy" comparisons and failing to adjudicate more ambiguous candidates, leading to a small statistical gain.

A more telling, albeit less numerically successful, result is from Qwen-VL. While its absolute Top-1 accuracy of 30.40\% remains low, it represents a \textit{relative} leap  compared to its best pointwise performance (25.00\%). This suggests that Qwen-VL, which already showed a (highly variant) signal for correct candidates in the pointwise tasks, is able to leverage the comparative prompt structure more effectively. This finding indicates that Qwen-VL's underlying capability is a promising candidate for this task, which could potentially be harnessed with fine-tuning or more advanced prompt engineering to surpass the baseline.
\begin{figure}[ht!] 
  \centering
  \includegraphics[width=1\columnwidth]{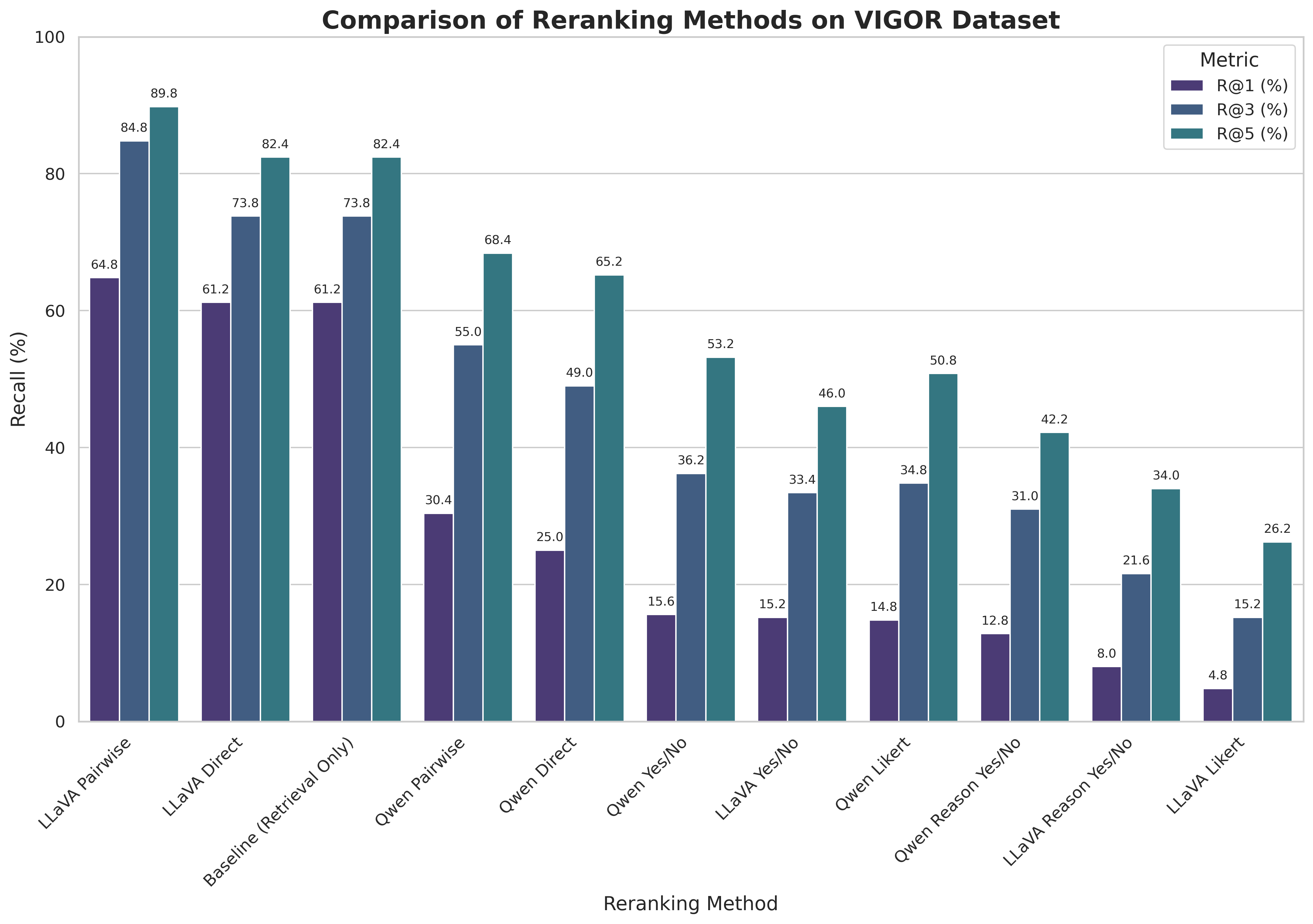} 
  \caption{Main Recall Performance Comparison. This plot shows Recall@1, Recall@3, and Recall@5 for the baseline retrieval model and all VLM reranking strategies, sorted by R@1 performance. LLaVA Pairwise (64.80\% R@1) is the only method to achieve a significant improvement over the Baseline (61.20\% R@1). In contrast, all other pointwise methods (Direct, Likert, Yes/No) cause a catastrophic drop in performance, or no change at all.}
  \label{fig:results_chart}
\end{figure}

\section{Conclusion}
\label{conclusion}

This work investigated the use of zero-shot Vision-Language Models to address the low top-1 accuracy of SOTA cross-view geolocalization systems. We proposed a two-stage framework, using a SOTA model for candidate retrieval and a VLM for subsequent reranking. We systematically compared pointwise (absolute scoring) and pairwise (relative comparison) strategies.

Our findings on the VIGOR dataset are twofold. First, all pointwise methods failed, with our analysis revealing that these VLMs appear poorly calibrated for assigning absolute similarity scores in this specific task configuration, as shown in Figure \ref{fig:point_plots}. Second, the \textbf{LLaVA Pairwise} strategy proved effective, improving the \textbf{Top-1 accuracy from 61.20\% to 64.80\%} and R@5 from 82.40\% to 89.80\%. 

We conclude that the \textit{task formulation} is critical: while VLMs fail at absolute scoring, they possess a strong emergent capability for fine-grained \textit{relative visual judgment}. This suggests that VLM-based pairwise reranking is a viable and promising direction for enhancing the precision of geolocalization systems. Future work should focus on fine-tuning VLMs on a dedicated pairwise cross-view dataset to further harness this capability.
{
	\begin{spacing}{1.17}
		\normalsize
		\bibliography{ISPRSguidelines_authors} 
	\end{spacing}
}
\appendix
\section{Appendix}
In this part, we will provide the prompts used in experiments and some qualitative results.
\label{appendix:prompts}
\subsection{Prompts Used In Experiments}

\vspace{-1mm} 
\begin{tcolorbox}[
  colback=purple!5!white, 
  colframe=purple!55!black, 
  title=Pointwise: Direct Prompt,
  top=2pt, 
  bottom=2pt 
]
\small 
You are an expert geospatial analyst. Your task is to determine the best satellite image match for a ground-level query panorama. Here is the ground-level panorama query image:

'$<$GROUNDIMAGE$>$'

Here is a candidate satellite image:

'$<$AERIALIMAGE$>$'

Carefully analyze the ground-level query image and compare it to satellite image. Focus on key cross-view features like road network, building arrangement, and landmarks. Evaluate if the satellite image corresponds to the location shown in the ground-level panorama. Provide a confidence score between 0 (no match) and 100 (perfect match). Respond ONLY with the score.
\end{tcolorbox}

\vspace{-1mm} 
\begin{tcolorbox}[
  colback=green!5!white, 
  colframe=green!50!black, 
  title=Pointwise: Yes/No Prompt,
  top=2pt, 
  bottom=2pt
]
\small
You are an expert geospatial analyst. Your task is to determine the best satellite image match for a ground-level query panorama.
Here is the ground-level panorama query image:

'$<$GROUNDIMAGE$>$'

Here is a candidate satellite image:

'$<$AERIALIMAGE$>$'

Carefully analyze the ground-level query image and compare it to satellite image. Focus on key cross-view features like road network, building arrangement, and landmarks. Based on this comparison does the satellite image accurately match the location shown in the ground-level panorama? Answer ONLY with the single word ‘Yes’ or ‘No’.
\end{tcolorbox}

\begin{tcolorbox}[
  colback=yellow!5!white, 
  colframe=yellow!85!black, 
  title=Pointwise: Likert Prompt,
  top=2pt, 
  bottom=2pt
]
\small
You are an expert geospatial analyst. Your task is to determine the best satellite image match for a ground-level query panorama. Here is the ground-level panorama query image:

'$<$GROUNDIMAGE$>$'

Here is a candidate satellite image:

'$<$AERIALIMAGE$>$'

Carefully analyze the ground-level query image and compare it to satellite image. Focus on key cross-view features like road network, building arrangement, and landmarks
On a scale of 1 (no match) to 5 (perfect match), how well does the satellite image match the location in the ground-level panorama? Respond ONLY with a single digit (1, 2, 3, 4, or 5).
\end{tcolorbox}

\begin{tcolorbox}[
  colback=purple!5!white, 
  colframe=purple!55!black, 
  title=Pointwise: Yes/No with Reasoning Prompt,
  top=2pt, 
  bottom=2pt
]
\small
You are an expert geospatial analyst. Your task is to determine the best satellite image match for a ground-level query panorama. Here is the ground-level panorama query image:

'$<$GROUNDIMAGE$>$'

Here is a candidate satellite image:

'$<$AERIALIMAGE$>$'

Carefully analyze the ground-level query image and compare it to satellite image. Focus on key cross-view features like road network, building arrangement, and landmarks. Provide a detailed step-by-step reasoning comparing key visual features (e.g., road layout, building shapes, relative positions, landmarks). Explain your conclusion. Based on the reasoning about the match between the two images. Concisely, does the satellite image match the ground-level panorama? Answer ONLY with the single word 'Yes' or 'No'.
\end{tcolorbox}

\begin{tcolorbox}[
  colback=blue!5!white, 
  colframe=blue!34!black, 
  title=Pairwise Prompt,
  top=2pt, 
  bottom=2pt
]
\small
You are an expert geospatial analyst. Your task is to determine the best satellite image match for a ground-level query panorama. Here is the ground-level panorama query image

'$<$GROUNDIMAGE$>$'

Here is satellite image 1:

'$<$AERIALIMAGE1$>$'

Here is satellite image 2:

'$<$AERIALIMAGE2$>$'

Carefully analyze the ground-level query image and compare it to both candidate satellite images. Focus on key cross-view features like road network, building arrangement, and landmarks Based on this comparison, which satellite image (1 or 2) provides the better geospatial match?
Respond ONLY with a JSON object indicating the preferred image number (1 or 2), like this: $\{$"preference": "$<$1 or 2$>$"$\}$
\end{tcolorbox}

\begin{figure*}[t!]
    \centering
    \begin{minipage}[t]{0.49\linewidth}
        \centering
        \includegraphics[width=\linewidth]{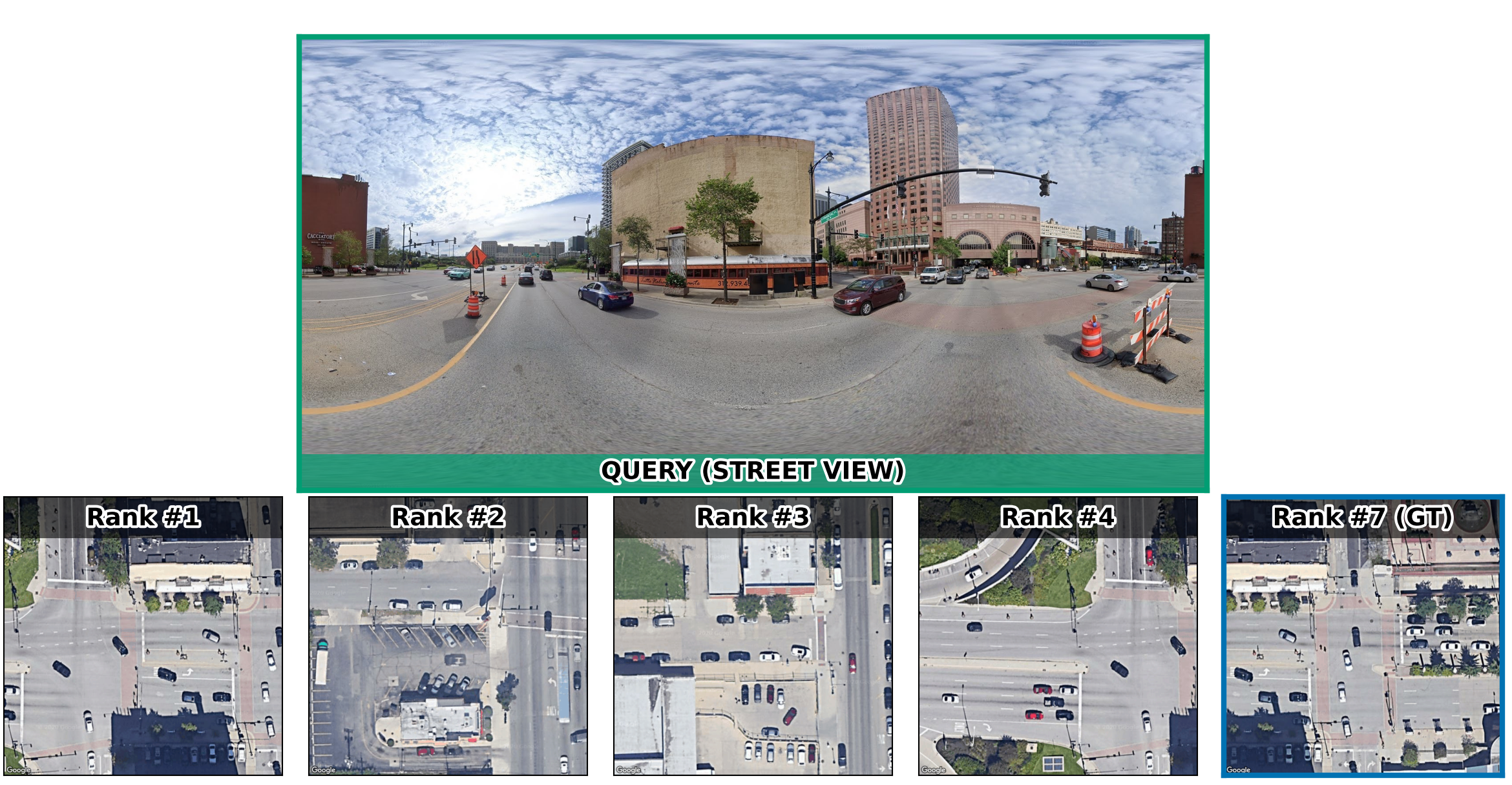}
    \end{minipage}
    \hfill
    \begin{minipage}[t]{0.49\linewidth}
        \centering
        \includegraphics[width=\linewidth]{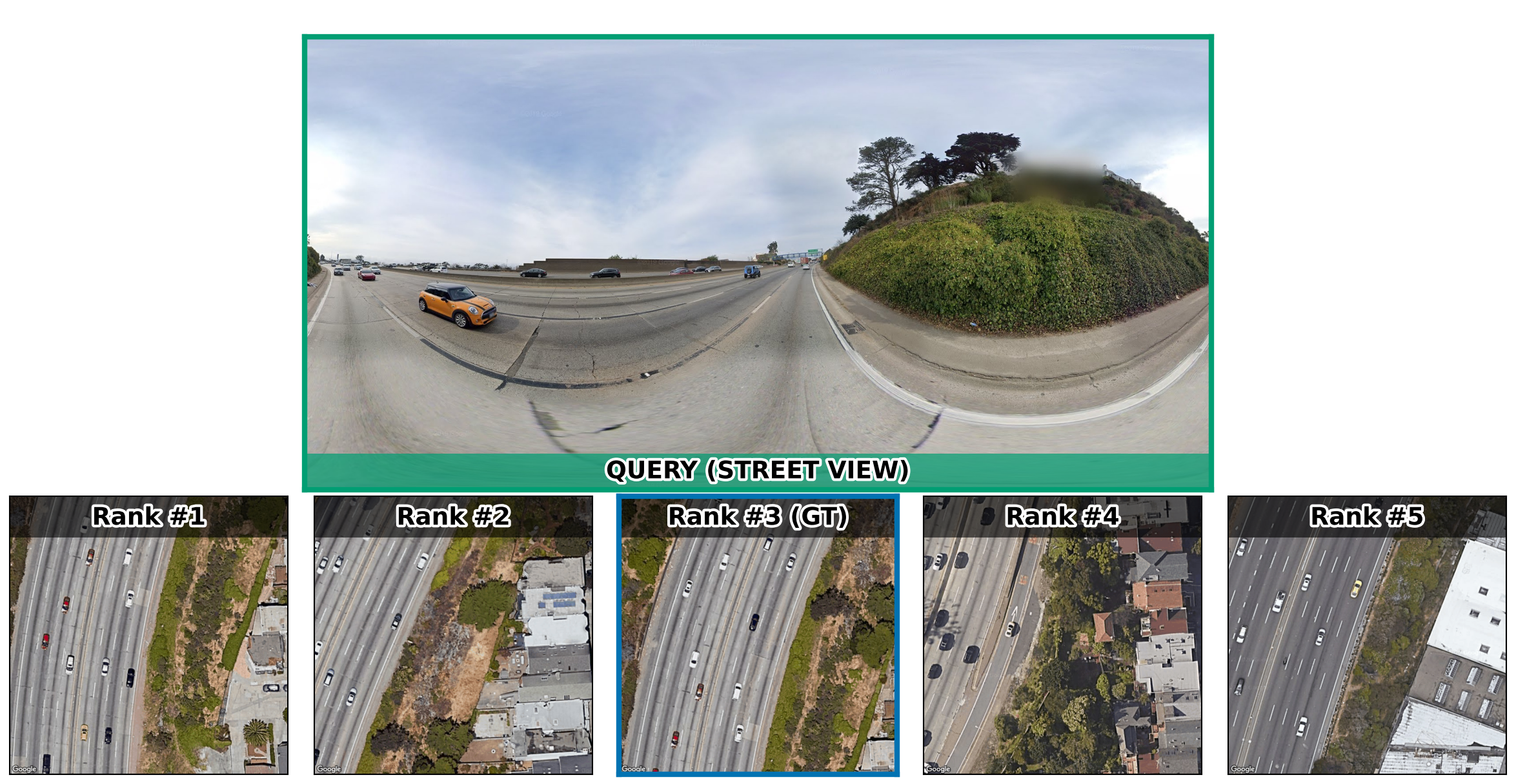}
    \end{minipage}

    \vspace{0.5em}

    \begin{minipage}[t]{0.49\linewidth}
        \centering
        \includegraphics[width=\linewidth]{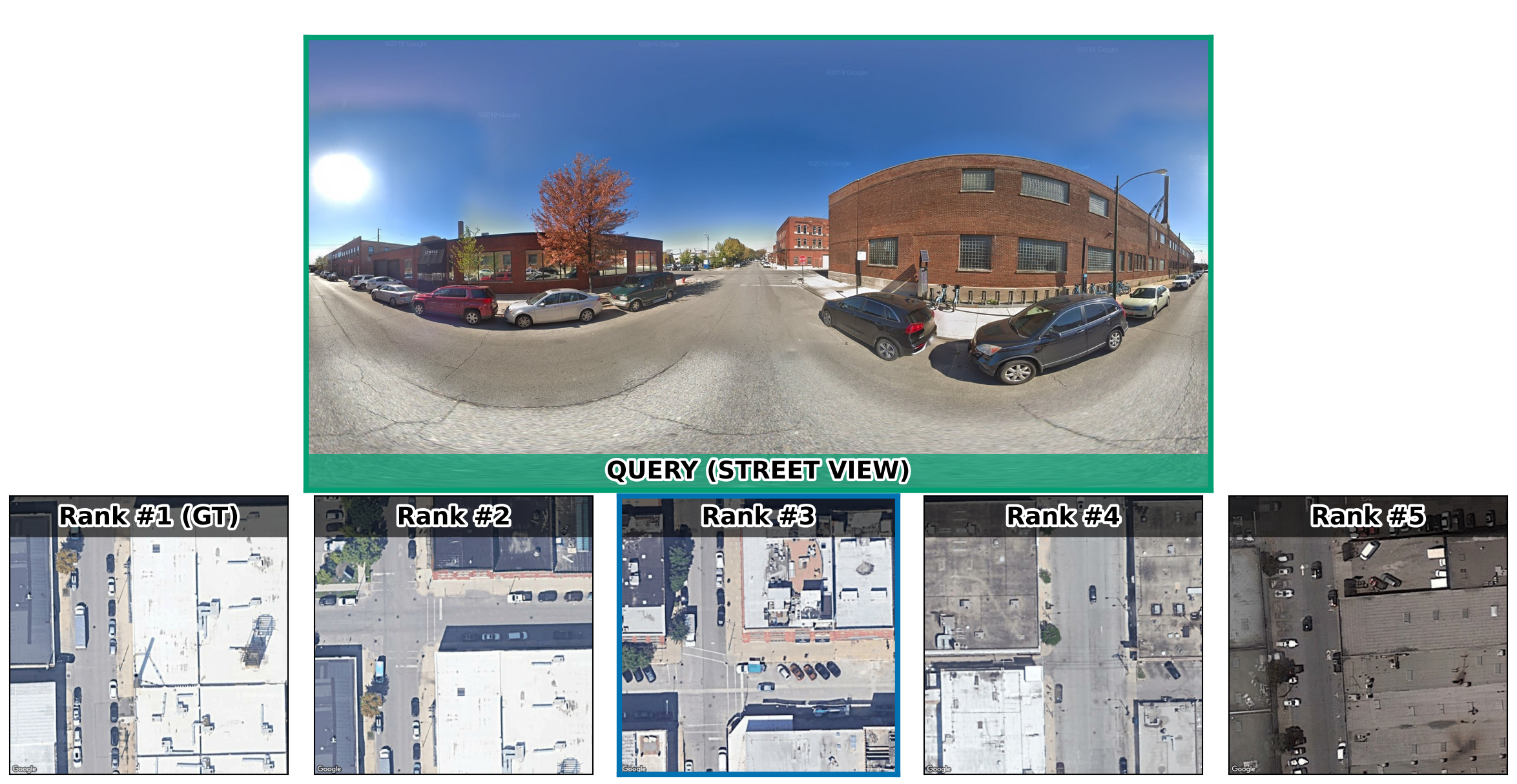}
    \end{minipage}
    \hfill
    \begin{minipage}[t]{0.49\linewidth}
        \centering
        \includegraphics[width=\linewidth]{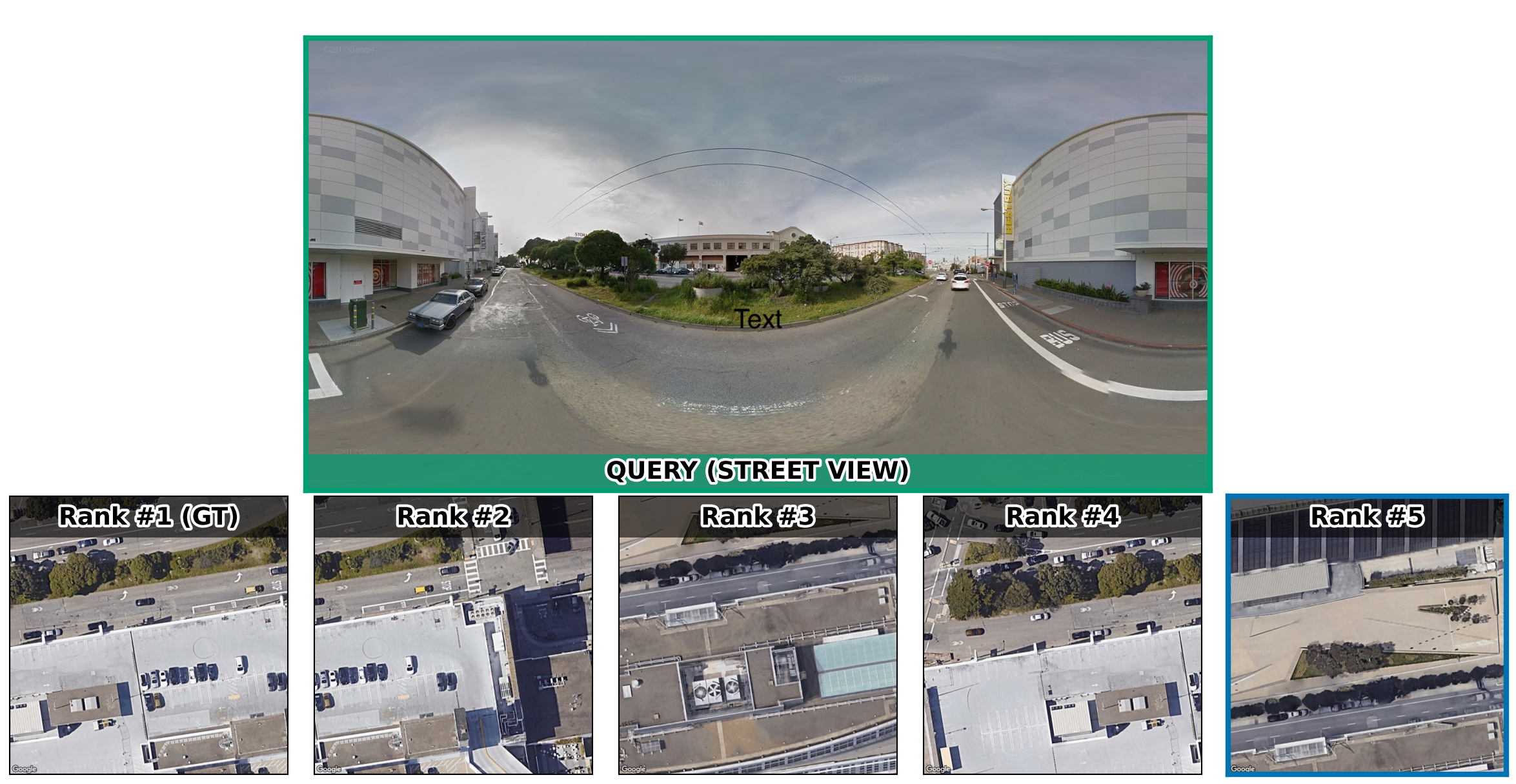}
    \end{minipage}
    \caption{Some qualitative results comparing the base model's ranking and the VLM's final selection. In each example, the satellite images are displayed in the order originally produced by the base model, \emph{not} in the reranked order. The blue outlined image indicates the VLM's top-1 selection among these candidates.}
    \label{fig:qualitative_results}
\end{figure*}
\subsection{Qualitative Results}
We provide several qualitative examples to illustrate how the VLM selects among the candidates proposed by the base model in Figure \ref{fig:qualitative_results}. In all examples, the satellite images are shown in the original order produced by the base model, while the blue outline marks the VLM's top-1 choice.
\end{document}